\documentclass[letterpaper]{article} 
\usepackage{amsmath}
\usepackage{aaai25}  
\usepackage{times}  
\usepackage{helvet}  
\usepackage{courier}  
\usepackage[hyphens]{url}  
\usepackage{breakurl}      

\usepackage{graphicx} 
\usepackage{natbib}  
\usepackage{caption} 
\usepackage{algorithm}
\usepackage{algpseudocode}

\usepackage{multirow}
\usepackage{bbding}
\usepackage{booktabs}
\usepackage{amssymb,amsmath,bm}
\usepackage{newfloat}
\usepackage{listings}
\DeclareCaptionStyle{ruled}{labelfont=normalfont,labelsep=colon,strut=off} 
\floatstyle{ruled}
\newfloat{listing}{tb}{lst}{}
\floatname{listing}{Listing}
\title{ELDER: Enhancing Lifelong Model Editing with Mixture-of-LoRA}
\author{
    Jiaang Li$^{\dagger}$, Quan Wang$^{\ddagger}$, Zhongnan Wang$^{\dagger}$, Yongdong Zhang$^{\dagger}$, Zhendong Mao$^{\dagger}$\thanks{~~Corresponding author: Zhendong Mao}
}
\affiliations{
    $^{\dagger}$ University of Science and Technology of China\\
$^{\ddagger}$Beijing University of Posts and Telecommunications
 
    jali@mail.ustc.edu.cn, wangquan@bupt.edu.cn \{zhyd73, zdmao\}@ustc.edu.cn
}

\usepackage{bibentry}

\begin{document}

\maketitle

\begin{abstract}
Large language models (LLMs) require model editing to efficiently update specific knowledge within them and avoid factual errors. 
Most model editing methods are solely designed for single-time use and result in a significant forgetting effect in lifelong editing scenarios, where sequential edits are conducted over time.
Previous approaches manage sequential edits by freezing original parameters and discretely allocating new parameters for each knowledge update.
However, these methods lack robustness to minor input variations due to the discrete mapping between data and parameters. 
To overcome this challenge, we propose ELDER, a novel approach to create a continuous association between data and adapters. 
ELDER integrates multiple LoRAs through a router network and is trained to establish a smooth data-adapter association, thereby enhancing the edit robustness and generalization of semantically equivalent inputs.
To ensure inputs containing the same knowledge will be processed by the same LoRAs, we design a novel loss to guide the model link LoRA allocations with edit knowledge.
Furthermore, we propose a deferral mechanism to retain the original LLM capabilities post-edit. 
Extensive experiments on GPT-2 XL and LLaMA2-7B demonstrate that ELDER effectively edits models in the lifelong setting, outperforming eight baselines while exhibiting strong scalability and preserving LLMs' general abilities on downstream tasks. 
Our code is available at \burl{https://github.com/JiaangL/ELDER}. 
\end{abstract}

%

\section{Introduction}
Large language models (LLMs) are renowned for their text understanding and generation capabilities  (\citealp[]{gpt3}; \citealp[]{gpt4}; \citealp[]{touvron2023llama2}; \citealp[]{radford2019gpt2}). 
Despite their widespread use, LLMs often produce factual errors, including hallucinations and outdated information  (\citealp[]{ji2023towards}; \citealp[]{wang2023hallucination}; \citealp[]{tam2023evaluating}).
Retraining or fine-tuning to update the model is expensive and time-consuming. 
Therefore, model editing techniques, which modify specific knowledge within LLMs with low resources, are gaining increasing attention (\citealp[]{yao2023editing}; \citealp[]{mitchell2022memory_serac}; \citealp[]{meng2022locating_rome}).
In practice, evolving world knowledge necessitates repeated model edits over time, which is known as lifelong model editing  (\citealp[]{hartvigsen2024aging}; \citealp[]{yu2024melo}).

The most intuitive way to implement lifelong editing is to perform model editing methods successively for multiple times.
However, most model editing methods are designed for one-time use (\citealp[]{meng2022locating_rome}; \citealp[]{mitchell2022memory_serac}). 
Repeated using them causes LLMs to forget previous edits and pre-training data, significantly reducing their edit reliability and general ability on downstream tasks (\citealp[]{yao2023editing}; \citealp[]{gu2024model_hurt}; \citealp[]{yang2024butterfly}; \citealp[]{gupta2024model}; \citealp[]{lin2024navigating}; \citealp[]{huang2022transformer_patcher}).
Recently, a class of representative methods has been developed specifically for lifelong model editing (\citealp[]{hartvigsen2024aging}; \citealp[]{yu2024melo}). 
These methods freeze the original LLM parameters and incorporate additional adapters to modify the model. 
They cluster the input data and assign a specific adapter to each cluster, maintaining discrete data-adapter mappings. 
This approach enables the model to manage different knowledge with independent parameters, preventing interference between different edits. 
As a result, it ensures high reliability after sequential edits and outperforms other techniques.

However, these key-value mapping methods exhibit poor robustness and struggle with semantically equivalent inputs (\citealp[]{tian2024instructedit}; \citealp[]{lin2024navigating}). 
Due to the inherent discreteness of their data-adapter mappings, data points on opposite sides of a cluster boundary will map to entirely different adapters.  
Unfortunately, their clustering relies on manually set distance metrics and hyperparameters, resulting in inaccurate cluster boundaries. 
Semantically equivalent data with slight variations (e.g., rephrased sentences) could fall outside the appropriate cluster and are assigned incorrect adapters.
Consequently, these methods are not robust and prone to errors with rephrased edit data.

To address the robustness issues in previous discrete mapping methods, we propose a novel approach to create an adaptive and continuous association between factual knowledge and adapter parameters, named ELDER (\textbf{E}nhancing \textbf{L}ifelong mo\textbf{D}el \textbf{E}diting with mixtu\textbf{R}e-of-LoRA). 
ELDER successively updates model knowledge by utilizing a router network to integrate multiple LoRAs~\citep{hu2021lora}, akin to the Mixture-of-Experts (MoE) structure \citep{jacobs1991adaptive_moe_1, jordan1994hierarchical_moe_2}. 
Unlike previous lifelong model editing methods, it adaptively generates LoRA weights through end-to-end learning instead of manually-set distance metrics, and produces a weighted combination of top-\(k\) LoRA outputs. 
For different edits, ELDER dynamically adjusts LoRA weights to produce varied adapter allocations based on edit semantics, ensuring that semantically equivalent inputs are assigned similar allocations to generate consistent model responses. 
Thus, it ensures robust performance.  
Another advantage of ELDER lies in its scalability.
Unlike discrete mapping methods, which require new parameters for each modification, ELDER manages knowledge modifications seamlessly by various LoRA combinations rather than independent adapters.
Therefore, this approach avoids the need for additional parameters with each successive edit, allowing for scalability to longer editing sequences. 
To the best of our knowledge, this is the first work to employ a mixture-of-LoRA structure for model editing.

Beyond employing Mixture-of-LoRA, we also introduce two novel techniques designed to address specific challenges in the lifelong editing task. 
Specifically, we propose a guided loss function to align adapter allocation with edit knowledge and a deferral mechanism to retain general capabilities in post-edit models. 
Firstly, in the training data, semantically equivalent edits are preset with the same LoRA allocation. 
We design a novel training loss to guide the model in learning these allocations and establishing the association between LoRA allocation and data semantics, thus promoting the model to assign similar LoRA allocations to similar inputs during inference. 
Moreover, our deferral mechanism identifies whether an input requires editing based on its LoRA allocation. In this way, it concentrates on edit-related features while ignoring irrelevant details like input format to ensure accurate discrimination. 
For test inputs that differ significantly from edited samples, this mechanism deactivates the mixture-of-LoRA, retaining the model's original performance on downstream tasks while leveraging LoRAs for specific edits.

Through extensive experiments, we have demonstrated the effectiveness of ELDER. 
Our experiments are conducted on two popular LLMs, i.e., GPT2-XL\citep{radford2019gpt2} and LLaMA2-7B \citep{touvron2023llama2}, with two widely used model editing datasets, ZsRE \citep{levy2017zsre} and \uppercase {C}\textsc{ounter}\uppercase {F}\textsc{act} \citep{meng2022locating_rome}.
Results indicate that ELDER achieves better editing performance and enhances the robustness of rephrased edits by improving the editing generalization by over 10\% higher than eight baselines. It is also superior in reliably maintaining previous edits.
Furthermore, we show that ELDER retains most of post-editing LLM's abilities on downstream general tasks, significantly surpassing most existing methods.

\section{Related Works}
\label{Related Work}

\paragraph{Model Editing}
Model editing aims to accurately and efficiently modify knowledge in deep neural networks\citep{sinitsin2019editable}, especially language models\citep{yao2023editing}. 
Meta-learning-based methods MEND\citep{mitchell2021fast_mend} and KE\citep{de2021editing_ke} train an extra hypernetwork to learn changes in the base model.
KN\citep{dai2022knowledge_kn} attributes the neuron that embodies the knowledge and updates these neurons.
ROME\citep{meng2022locating_rome} locates the edit area by casual analysis and modifies the entire weight matrix.
SERAC\citep{mitchell2022memory_serac} trains a scope classifier to identify inputs that need to be edited, then uses a counterfactual model to process these inputs separately from the original model.

However, the above methods only consider static edits, i.e., modifying the model a single time.
\citet{huang2022transformer_patcher}
proposed a sequential editing approach by adding one neuron for each edit sample. Nevertheless, this method relies on large sets of unrelated inputs and is slow due to the need to train neurons for each edit   \citep{yao2023editing}. 
Most similar works to ours are GRACE \citep{hartvigsen2024aging} and MELO\citep{yu2024melo}, which discretely maps different adapters to successive edits.

\paragraph{Mixture-of-Experts (MoE)}
\citet{jacobs1991adaptive_moe_1}; \citet{jordan1994hierarchical_moe_2} introduced MoE models to compute different examples with independent expert modules. 
\citet{shazeer2016outrageously_moe_3} extended this concept to large-scale language models with LSTMs.
Advances such as GShard \citep{fedus2021switch}, Switch Transformer \citep{fedus2021switch}, and BASE Layer \citet{lewis2021base} have refined routing input tokens to experts. 
Hash Layer \citep{roller2021hash} uses a pre-defined token-level hash table for token-to-expert assignment. 
StableMoE \citep{dai2022stablemoe} addresses routing fluctuation by training the router network first and then freezing it for further model training. 
In contrast, our approach guides the router network to learn a pre-set sample-to-adapter assignment, effectively handling unseen but equivalent inputs.

\paragraph{Mixture-of-LoRAs}
LoRA \citep{hu2021lora} uses low-rank matrices to update the LLMs, and has become a popular parameter-efficient fine-tuning method.
Recently, researchers have been using combinations of multiple LoRAs for further benefits.
\citet{huang2023lorahub} proposed composing existing LoRA modules for generalization on new tasks.
\citet{zadouri2023pushing_loramoe_1} combined LoRA with token-level mixture-of-experts \citep{jacobs1991adaptive_moe_1} and designed a new PEFT module.
\citet{dou2023loramoe_2} split LoRA experts into two separate groups to maintain world knowledge of LLMs during instruction tuning. 
MoRAL \citep{yang2024moralmoeaugmentedlora} augments LoRA with MoE to achieve both multi-task and fine-tuning abilities.

Although sharing a similar base structure with previous mixture-of-LoRA research, our approach distinguishes itself from these works in following aspects. 
First, we focus on the task of editing factual knowledge within the model and demonstrating strong effectiveness, which previous works have not explored.
Second, our routing scheme is tailored for model editing. We route the whole sequence altogether to ensure equal treatment of knowledge, unlike previous works that apply different experts to individual tokens.
Moreover, we introduce novel techniques to enhance the adaptability of mixture-of-LoRA to the lifelong editing task, including the guided loss function and deferral mechanism.

\begin{figure*}
    \centering
    \setlength{\belowcaptionskip}{-0.15cm}
    \includegraphics[width=0.95\linewidth]{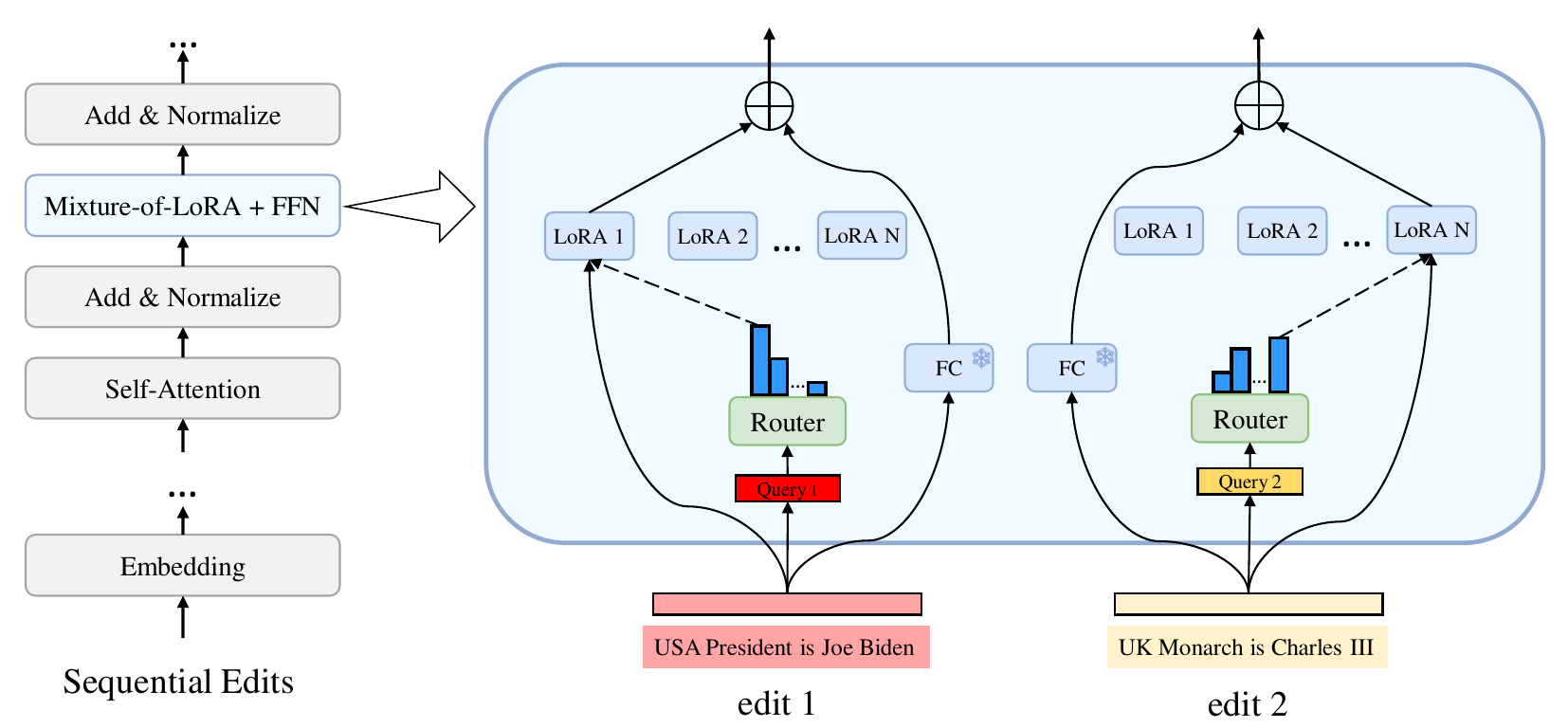}
    \caption{An illustration of processing two different edits with the mixture-of-LoRA module in ELDER. A mixture-of-LoRA module is applied to the FC layer at the FFN of the Transformer block. Each edit is routed to top-\(k\) LoRAs with the highest scores based on its query vector. This figure takes \(k=1\) as an example. The final results are summations of LoRA outputs and outputs of the original FC. Dotted lines denote multiplying LoRA outputs with corresponding weights. Training loss and deferral mechanism are omitted in this figure for simplicity.}
    \label{fig:method_model_structure}
\end{figure*}

\section{Proposed Methods}
\label{Methods}

Our proposed ELDER establishes a continuous and smooth connection between data and adapters.
It contains multiple mixture-of-LoRA modules, which adaptively allocate LoRAs to each successive edit.
Moreover, an auxiliary training loss is specially designed for lifelong model editing to assist the model in aligning LoRA allocations with the input knowledge. 
During inference, these LoRA allocations help to identify task inputs that do not require editing via a novel deferral mechanism tailored for the lifelong editing task. Such inputs are processed using the original LLM, thereby preserving the model's performance on general tasks. 

\subsection{Problem Formulation}
The lifelong model editing task, as described by \citet{hartvigsen2024aging}, involves continuously editing an initial base LLM, \(f_{base}\), without harming its overall abilities or negating the corrections made by prior edits.
Let \(D_{edit}=\{e_1,\dots,e_n\}\) denote \(n\) successive edits applied to the model, where each edit \(e_i=(x_i,y_i)\). 
The primary objective is to obtain a post-edit model \(f_n\) that correctly maps an input \(x\)  to its corresponding prediction \(y\), i.e., \(f_n(x_i)=y_i\). 
Furthermore, \(f_n\) should be capable of predicting the equivalent neighbor \(N(e_i)\) \citep{yao2023editing}, such as rephrased sentences, ensuring \(f_n(x_i')=y_i'\) for all \((x_i',y_i')\in N(e_i)\). 
Additionally, \(f_n\) must retain the performance on its original test data for practical usability\citep{hartvigsen2024aging}.
Given \(k\) example tasks \(t_1,\dots,t_k\) and their respective metrics \(m_1,\dots,m_k\), \(f_n\) should preserve \(f_{base}\)'s abilities across them, i.e., \(m_j(t_j, f_n)=m_j(t_j, f_{base}), \forall j\in \{1,\dots,k\}\).

\subsection{Mixture-of-LoRA Structure}
ELDER employs mixture-of-LoRA modules as the base structure.
Each mixture-of-LoRA module comprises \(N\) LoRAs and a router network, as shown in Figure \ref{fig:method_model_structure}. 
The router network, implemented as a fully connected (FC) layer, takes a text sequence's query vector as input and routes the entire sequence to the top-\(k\) LoRAs selected from a pool of \(N\) LoRAs.  
Notably, we route all tokens in the same instance to one LoRA allocation based on the query vector, ensuring equal treatment of the entire knowledge.
Following prior works \citep{hartvigsen2024aging,yu2024melo}, we define the query vector as the hidden representation of the last token in the input sequence, denoted as \(\textbf{x}\in \mathbb{R}^d\).
The router network generates scores \(\mathbf{s}(\mathbf{x})\) for each LoRA by normalizing the projection of \(\textbf{x}\) using a softmax distribution over the available \(N\) LoRAs at that layer, which are given by:
\[
\mathbf{s}(\mathbf{x}) = \text{softmax}(\mathbf{W}_r \cdot \mathbf{x}),
\]
where \(\mathbf{W}_r \in \mathbb{R}^{N \times d}\) is a projection matrix. 
This process adaptively captures the semantic association between edits and their rephrases, assigning them similar LoRA scores.

We further apply top-\(k\) gating \citep{shazeer2016outrageously_moe_3}, selecting \(k\) LoRAs with the highest scores for routing. 
Each selected LoRA module generates a product matrix \(\Delta\mathbf{W}_i=\mathbf{B}_i\mathbf{A}_i\), where \(\mathbf{B}_i\in\mathbb{R}^{d\times r}\) and \(\mathbf{A}_i\in\mathbb{R}^{r\times k}\) are two low-rank matrices in the \(i\)-th LoRA module.
The updated matrix \(\Delta \mathbf{W}\) is obtained by computing the linearly weighted combination of the selected LoRA matrices based on the score values, 
\[
\Delta \mathbf{W}=\sum_{i\in \mathcal{T}}s_i(\mathbf{x})\cdot \Delta \mathbf{W}_i,
\]
where \(s_i\) is the \(i\)-th element of \(\mathbf{s}\), and \(\mathcal{T}\) is the set of selected top-\(k\) indices.

We inject a mixture-of-LoRA module into the feed-forward network (FFN) of the Transformer block to edit the original FC layer, as shown in Figure \ref{fig:method_model_structure}. 
The forward pass of a standard FC layer is formulated as
\[\mathbf{y}=\mathbf{W}_{0}\mathbf{v}+\mathbf{b},\]
where \(\mathbf{b}\in\mathcal{R}^k\) is the bias, \(\mathbf{W}_{0}\in\mathbb{R}^{d\times k}\) is the weight matrix, and \(\mathbf{y}\) and \(\mathbf{v}\) are the output and input vectors, respectively.
The modified FC computation involves multiplying both \(\mathbf{W}_{0}\) and \(\Delta\mathbf{W}\) with the input \( \mathbf{v}\) and summing the result coordinate-wise. 
The modified forward pass is:
\[\mathbf{y}=\mathbf{W}_{0}\mathbf{v}+\Delta\mathbf{W}\mathbf{v}+\mathbf{b}.\]
Here, \(\mathbf{W}_{0}\) is the FC layer's original weight matrix, which remains frozen during training. The modified FC computation is applied to all tokens in the given input sequence. 
Furthermore, as illustrated in the figure, ELDER can handle more edits through adapter combinations without requiring additional learnable parameters for each new edit. 
In contrast, previous discrete mapping methods increase the parameter count linearly for each new edit, implying scalability issues.

\subsection{Guided Loss}\label{sec: guide}
We propose a novel training loss, guiding the model to allocate LoRAs based on knowledge embedded in each input. 
This training objective promotes the model to associate input semantics with various adapter allocations, thereby cultivating the adaptability to assign similar allocations to semantically equivalent inputs during inference. 
Specifically, we first pre-assign a LoRA allocation for each edit in the training data via random generation, ensuring that identically labeled edits receive the same allocation distinct from others. 
Subsequently, our proposed training loss guides the router network in learning to assign these allocations.

Formally, we guide the model with a new loss, \(L_{guide}\), by maximizing the probability of selecting the pre-assigned LoRAs.
Suppose the model employs mixture-of-LoRA to  \(L\) layers. 
Each produced allocation \(\mathcal{A}\) should contain \(L\times k\) indices, indicating selected adapters from a total \(L\times N\) available LoRAs.
Therefore, the loss function takes the form of:
\[
\begin{aligned}
\mathcal{L}_{guide}= \sum_{i,j\in\mathcal{A}}-\log(s_{i,j}),\\ 
\mathcal{L}_{total}=\mathcal{L}_{model}+\lambda\mathcal{L}_{guide}, 
\end{aligned}
\]
where \(s_{i,j}\) is the score of the chosen \(j\)-th LoRA at the \(i\)-th layer. 
\(L_{total}\) is the total loss to be optimized, \(L_{model}\) is the original model loss, and \(\lambda\) is a hyperparameter. 
Intuitively, this guiding process encourages the model to generate a unique allocation for each piece of knowledge, thereby avoiding interference between different knowledge and improving parameter utilization.

Although previous MoE methods \citep{fedus2021switch, dai2022stablemoe} improve the token-routing scheme by balancing the usage load of parallel modules to achieve uniform distribution, they present a challenge in lifelong model editing. 
In this context, sequential edits occur sparsely, and all tokens in a single edit apply the same routing scheme rather than being routed separately, as they represent the same knowledge. Thus, each training batch has fewer samples, making it difficult to calculate the average usage load accurately. 
This disparity means that batch measurements do not accurately reflect real usage loads, rendering the optimization for a uniform distribution inapplicable to our task.

\subsection{Deferral Mechanism}

\begin{algorithm}[t]
    \caption{Inference with deferral mechanism.}
    \label{algorithm}

    \begin{algorithmic}[1]
        \Statex \textbf{Input:} Test input $x$. Threshold \(\epsilon\)
        \Statex \textbf{Input:} Model \(M\) with layers \(\{M_i\}_{i=1}^t\). First mixture-of-LoRA layer is \(M_{l_0}\)
        \Statex \textbf{Input:} Allocation codes of all edits \(\{\mathbf{c}_e^i\}_{i=1}^n\)
        \Statex \textbf{Output:} Model response \(R\)
        
        \State $layer\_in \leftarrow x$
        \State $flag \leftarrow 0$
        
        \For{\(l\) in \(1\) to \(t\)}
            \If{\(l = l_0 - 1\)}
                \State \(layer\_out \leftarrow M_l(layer\_in, flag)\)
                \State \(\mathbf{c} \leftarrow GetAlloc(M, layer\_out)\) \Comment{Get allocation code with all router networks.}
                \State \(dist \leftarrow \min_{i} HamDist(\mathbf{c}, \mathbf{c}_e^i)\) \Comment{Find the nearest distance.}
                \If{\(dist < \epsilon\)}
                    \State \(flag \leftarrow 1\) \Comment{Following layers will use mixture-of-LoRAs if available.}
                \Else
                    \State \(flag \leftarrow 0\) \Comment{Use original model parameters.}
                \EndIf
            \Else
                \State \(layer\_out \leftarrow M_l(layer\_in, flag)\)
            \EndIf

        \EndFor
        
        \State \(R \leftarrow layer\_out\)
    \end{algorithmic}
\end{algorithm}

During inference, we aim to preserve the original capabilities of LLMs on general tasks to ensure their practical application. To this end, we design a novel deferral mechanism to identify inputs that do not involve edited knowledge, referred to as task inputs. 
The proposed mechanism is based on the LoRA allocations, since they contain edit-related features within the input, excluding the impact of irrelevant details like input format as described in Section \ref{sec: guide}. 
After distinguishing these non-edited inputs, we deactivate the mixture-of-LoRA module and process them directly with the original model. This mechanism ensures that task inputs yield the same output as they would from the initial model. 

The detailed steps of this deferral mechanism are outlined in Algorithm \ref{algorithm} and described next.
First, test input is compared to all edited samples via the allocation code, a \(L\times N\) boolean vector with \(L\times k\) nonzero elements indicating the top-\(k\) LoRAs at each layer. 
Formally, for a test input with allocation \(\mathcal{A}\), its allocation code \(\mathbf{c}\) is defined as:
\[
c_{i\times N + j}=\begin{cases}
1 & \text{if} \quad (i,j)\in\mathcal{A}, \\
0 & \text{if} \quad (i,j)\notin\mathcal{A},
\end{cases}
\]
where \((i,j)\) denotes the indices of the \(j\)-th LoRA at the \(i\)-th mixture-of-LoRA layer.
\(\mathbf{c}\) is represented by boolean values, and computed using all router networks before the first mixture-of-LoRA (line 6)  to discriminate model inputs in advance for efficient editing.
The allocation codes \(\{\mathbf{c}_e^i\}_{i=1}^n\) for all edits are precomputed and stored during the editing stage
The nearest distance between \(\mathbf{c}\) and all items in \(\{\mathbf{c}_e^i\}_{i=1}^n\) is then calculated using the Hamming distance (line 7). 
This distance computation is highly efficient because all allocation codes are boolean, allowing the operations between them to be performed using simple bitwise operations. 
Inputs whose nearest distance exceeds a threshold \(\epsilon\) are identified as task inputs and are processed with the preserved original parameters (line 9).

\section{Experiments}
\subsection{Experimental Setup}
\subsubsection{Baselines}
We compare our proposed method, ELDER, with eight baselines. 
We first select two methods tailored for the lifelong editing setup. 
They preserve the original model and modify it with discrete data-adapter mappings.
\textbf{GRACE} \citep{hartvigsen2024aging} uses representation vectors as adapters, writing them into the pre-trained model's latent space based on specific data samples.
\textbf{MELO} \citep{yu2024melo} builds on the same foundation framework but uses LoRAs as adapters. 
Other baselines include \textbf{SERAC} \citep{mitchell2022memory_serac}, which additionally trains a scope classifier and a counterfactual model to alter the model behavior, and \textbf{T-Patcher} \citep{huang2022transformer_patcher}, which adjusts one neuron for each new edit. 
We also include \textbf{ROME} \citep{meng2022locating_rome}, a state-of-the-art one-time model editing method. 
It locates the edit area in GPTs via casual tracing and updates relevant weights. 
We further include \textbf{WilKE}\citep{hu2024wilke}, which modifies different layers for different edits.
Beyond these model editing techniques, we compare ELDER with \textbf{FT-L} \citep{meng2022locating_rome}, which fine-tunes the layers identified by ROME, and with the standard \textbf{LoRA} \citep{hu2021lora}.

\subsubsection{Metrics}
We apply three metrics to evaluate both the editing performance and the preservation of post-edit model capability. These metrics align with those reported in previous works \citep{meng2022locating_rome,gu2024model_hurt, yao2023editing}.

\subparagraph{Reliability}
An edit \(e_i=(x_i, y_i)\) is reliable if the post-edit model \(f_n\) generates the target answer correctly. 
We check how well \(f_n\) retains previous edits by reliability, which is measured as the average accuracy of the edit data:
\[
\mathbb{E}_{e_i\in D_{edit}}\mathbb{I}\{\operatorname*{argmax}_y f_n(y|x_i)=y_i\}.
\]
\subparagraph{Generalization}
We check how well \(f_n\) generalizes to semantic equivalent data \(N(e_i)\), e.g. rephrased sentences, by the average accuracy on these data:
\[
\mathbb{E}_{(x_i',y_i')\in N(e_i), e_i\in D_{edit}}\mathbb{I}\{\operatorname*{argmax}_y f_n(y|x_i')=y_i'\}.
\]
\subparagraph{Test Retention on General Tasks}
We aim to evaluate the side effects of model editing on the capabilities of LLMs. Although a previous metric, \textit{Locality}, is designed for this purpose, it is limited in scope, focusing on a narrow data range and simple QA task, which is insufficient to assess the full functionality of LLMs, which can lead to inflated results and fails to assess the full functionality of LLMs  \citep{yang2024butterfly, gupta2024model}. 
To achieve a more comprehensive evaluation,  we extend the assessment to a broader range of tasks, following the approach of \citet{gu2024model_hurt}.
Specifically, we measure how well  \(f_n\) retains its original performance during inference across \(k\) representative general tasks. 
The average result is then used to determine its retention on these tasks: 
\[
\frac{1}{k}\sum\limits_{j=1}^{k} m_j(t_j,f_n).
\]

\subsubsection{Datasets}\label{sec: dataset}
We evaluate the reliability and generalization of ELDER on lifelong model editing with two widely used model editing datasets for training and evaluation.
The first one, ZsRE \citep{levy2017zsre}, is a zero-shot relation extraction dataset that uses question rephrasing generated through back-translation as the rephrased edit data. 
The second dataset, \uppercase {C}\textsc{ounter}\uppercase {F}\textsc{act} \citep{meng2022locating_rome}, is a more challenging dataset.
It comprises counterfactual statements initially receiving low factuality scores. 
We adopt both datasets to the lifelong model editing setting by extracting a sequence of 1000 editing samples with their rephrasings for our main experiments, following the methodologies outlined in \citep{hartvigsen2024aging} and \citep{yu2024melo}. Further details are in the technical appendix.

\begin{table*}[htbp]
  \centering
\setlength\tabcolsep{1.5pt}
                    \begin{tabular}{ccc|c|cccccccc|c}
    \toprule
    \multirow{2}[4]{*}{Dataset} & \multirow{2}[4]{*}{Model} & \multicolumn{1}{c|}{\multirow{2}[4]{*}{Metric}} & \multicolumn{10}{c}{Method} \\
\cmidrule{4-13}         &      &      & Base & FT-L & LoRA & ROME & WilKE & SERAC & T-Patcher & MELO & GRACE & ELDER \\
    \midrule
    \multirow{4}[4]{*}{ZsRE} & \multirow{2}[2]{*}{GPT2-XL} & Reliability & 0.00  & 48.23  & 30.22  & 48.40  & 53.32  & 96.09  & 77.29  & 70.81  & 96.80  & \textbf{97.47 } \\
         &      & Generalization & 0.00  & 47.61  & 19.39  & 47.20  & 47.10  & 53.03  & 67.74  & 66.41  & 0.00  & \textbf{96.08 } \\
\cmidrule{2-13}         & \multirow{2}[2]{*}{LLaMA2-7B} & Reliability & 0.25  & 32.21  & 10.70  & 77.60  & 61.84  & 83.14  & 62.94  & 64.57  & 89.48  & \textbf{93.96 } \\
         &      & Generalization & 0.37  & 28.96  & 7.31  & 74.53  & 51.29  & 60.38  & 48.37  & 42.93 & 0.46  & \textbf{90.21 } \\
    \midrule
    \multirow{4}[4]{*}{CounterFact} & \multirow{2}[2]{*}{GPT2-XL} & Reliability & 0.00  & 55.10  & 37.21  & 69.00  & 68.19  & \textbf{100.00 } & 91.36  & 65.80  & 88.90  & 94.65  \\
         &      & Generalization & 0.00  & 44.06  & 34.55  & 68.74  & 53.35  & 79.49  & 80.31  & 49.20  & 76.05  & \textbf{91.26 } \\
\cmidrule{2-13}         & \multirow{2}[2]{*}{LLaMA2-7B} & Reliability & 0.40  & 66.26  & 10.16  & 79.37  & 74.78  & 82.67  & 88.93  & 51.39  & 77.70  & \textbf{95.07 } \\
         &      & Generalization & 0.24  & 44.65  & 5.21  & 80.30  & 55.83  & 71.83  & 77.75  & 35.71  & 63.35  & \textbf{90.79 } \\
    \bottomrule
    \end{tabular}%
      \caption{Lifelong model editing performance of ELDER and baselines. 'Base' denotes the pre-editing models. All metrics shown are computed after all sequential edits, and higher is better. The best results are bolded.}
  \label{tab:main}%
\end{table*}%

To evaluate the test retention of post-edit LLMs on general tasks, we expand the evaluation task data used in previous studies \citep{hartvigsen2024aging,yu2024melo} for a more comprehensive assessment of the post-edit LLMs' general abilities. 
This expansion is crucial because recent studies show that current model editing methods can degrade LLM performance on downstream tasks \citep{gu2024model_hurt,gupta2024model,yang2024butterfly}, underscoring the importance of assessing their side effects with diverse datasets.
Specifically, we employ a benchmark from \citep{gu2024model_hurt}, including eight diverse tasks:
\textbf{Reasoning} on GSM8K \citep{cobbe2021training_gsm8k}, 
\textbf{Natural Language Inference} on RTE\citep{dagan2005_rte}, 
\textbf{Open-domain QA} on Natural Question\citep{kwiatkowski2019natural_NQ}, 
\textbf{Closed-domain QA} on BoolQ\citep{clark2019boolq}, 
\textbf{Dialogue} on MuTual\citep{cui2020mutual}, 
\textbf{Summarization} on SAMSum \citep{cui2020mutual}, 
\textbf{Named Entity Recognition} on CoNLL03\citep{sang2003conll}, and
\textbf{Sentiment Analysis} on SST2\citep{socher2013sst}. 
Their respective metrics are in the technical appendix.


\subsubsection{Implement Details}
We use two LLMs as base models: LLaMA2 (7B) \citep{touvron2023llama2} and GPT2-XL (1.5B)\citep{radford2019gpt2}.
To evaluate our baselines, we use their original implementations and adapt them to our datasets and base models.
For our proposed ELDER across all settings, the rank of LoRAs is set to 8, and the number of layers that apply mixture-of-LoRA is set to 6. 
The number of LoRAs per layer is set to 4, \(k\) is set to 2, and \(\epsilon\) is set to 12. \(\lambda\) is set to \(1e-2\). 
More details of training and hyperparameter tuning are available in the technical appendix.

\subsection{Experimental Results}
\subsubsection{Main Results}
We compare the lifelong model editing performance of our proposed ELDER with recently advanced baselines, as shown in Table \ref{tab:main}. 
Initially, the base models show low performance, indicating that the edit knowledge is not inherent to the model. 
After long sequences of edits, we observe that ELDER consistently outperforms these baselines, across all datasets and base models. 
FT-L and ROME, which are not designed for sequential editing, tend to forget previous edits.
GRACE, which relies on discrete data-adapter mapping, is a strong baseline for reliably remembering previous edits but lacks robustness when handling semantically equivalent inputs, leading to poor generalization scores, especially on ZsRE, as noted in previous works \citep{tian2024instructedit,lin2024navigating}.  
SERAC and T-Patcher also exhibit good lifelong editing reliability, but fall short in generalization.
In contrast, ELDER excels in maintaining previous edits and robustly generalizes to their rephrases, demonstrating its superiority.

\begin{table}[t]
\setlength\tabcolsep{2.5pt}
\renewcommand{\arraystretch}{1} 
  \centering
  \setlength{\belowcaptionskip}{-0.6cm}
  
                \begin{tabular}{lcccccccrc}
    \toprule
         & \multicolumn{3}{c}{ZsRE} &      & \multicolumn{3}{c}{\uppercase {C}\textsc{ounter}\uppercase {F}\textsc{act}} &      & \multirow{2}[4]{*}{\textit{Avg.}} \\
\cmidrule{2-4}\cmidrule{6-8}         & GPT2 &      & LLaMA2 &      & GPT2 &      & LLaMA2 &      &  \\
    \midrule
    Base & 30.2  &      & 32.1  &      & 30.2  &      & 32.1  &      & 31.2  \\
    \midrule
    FT-L & 0.7  &      & 25.2  &      & 0.7  &      & 7.8  &      & 8.6  \\
    LoRA & 0.2  &      & 5.9  &      & 0.2  &      & 0.6  &      & 1.7  \\
    ROME & 0.2  &      & 6.7  &      & 0.2  &      & 0.5  &      & 1.9  \\
    WilKE & 2.1  &      & 7.8  &      & 1.2  &      & 5.3  &      & 4.1  \\
    SERAC & 30.2  &      & 32.0  &      & 30.2  &      & 32.1  &      & 31.1  \\
    T-Patcher & 2.7  &      & 1.1  &      & 0.1  &      & 0.9  &      & 1.2  \\
    MELO & 29.6  &      & 29.9  &      & \textbf{31.4 } &      & 31.9  &      & 30.7  \\
    GRACE & \textbf{30.3}&      & \textbf{32.7 } &      & 30.3  &      & \textbf{32.2 } &      & \textbf{31.4 } \\
    \midrule
    ELDER & 30.1  &      & 32.3  &      & 30.5  &      & 31.5  &      & 31.1  \\
    \bottomrule
    \end{tabular}%

\caption{Test retention on general tasks of post-edit LLMs after successive edits from ZsRE and \uppercase {C}\textsc{ounter}\uppercase {F}\textsc{act}.
'Base' represents the original model performance. The highest numbers are bolded.}
\label{tab:General Ability}
\end{table}%

 Table \ref{tab:General Ability} reports the test retention on general tasks after sequential edits by each method. 
 Detailed results for each task are provided in the technical appendix. 
ELDER effectively retains the LLM general abilities after lifelong editing, preserving the practical value of post-edit models. 
The results demonstrate that our deferral mechanism successfully identifies task inputs using edit-related information from LoRA allocations.  
On the other hand, FT-L, LoRA,  ROME, WilKE and T-Patcher significantly degrade LLM performance due to repeated modifications of model parameters, consistent with previous studies \citep{gu2024model_hurt, gupta2024model}.

\subsubsection{Editing Efficiency and Scalability}
Efficiency and scalability are critical for lifelong model editing methods, as they enable the model to manage multiple sequential edits with acceptable cost, when the edit sequence length continually increases. 
We compare the editing efficiency of ELDER against GRACE and SERAC in Table \ref{tab:efficiency}. 
Both methods exhibit strong performance in lifelong model editing and the retention of general abilities.
We show the number of extra parameters and average editing time after 1000 sequential edits from the ZsRE dataset. 
The results clearly indicate that ELDER is more efficient than the baseline method. 
ELDER's time efficiency stems from its end-to-end design, which avoids the need to search for discrete mappings compared with GRACE. 
The editing speed of SERAC is not reported, because it requires additional training of the scope classifier and counterfactual model for each new edit. 
Moreover, we show the inference efficiency of ELDER in the technical appendix.

Furthermore, we extend the sequential edits from \(1000\) to \(4000\) to investigate the editing scalability. 
Figure \ref{fig:change_sequence_length_grace} illustrates the editing reliability and parameter amounts after different numbers of edits from ZsRE using LLaMA2-7B, averaged over five seeds. 
Notably, while GRACE requires increasing parameters with more edits, ELDER maintains high performance with a fixed parameter count, effectively accommodating more edits. This strong scalability of ELDER is attributed to its mixture-of-LoRA structure, which combines existing adapters to handle new edits rather than introducing independent parameters for each edit. 

\begin{table}[t]
  \centering
  \renewcommand{\arraystretch}{0.95} 
  \setlength{\belowcaptionskip}{-0.2cm}
\setlength\tabcolsep{1pt}
            \begin{tabular}{lccccc}
    \toprule
         & \multicolumn{2}{c}{GPT2-XL} &      & \multicolumn{2}{c}{LLaMA2-7B} \\
\cmidrule{2-3}\cmidrule{5-6}         & Speed & \textit{\#Param} &      & Speed & \textit{\#Param} \\
    \midrule
    SERAC & \textbackslash{} & 181M &      & \textbackslash{} & 224M \\
    GRACE & 13.56 s/edits & 6.4M &      & 7.47 s/edits & 4.1M \\
    ELDER & 1.82 s/edits & 3.2M &      & 2.12 s/edits & 1.6M \\
    \bottomrule
    \end{tabular}%
    \caption{Editing efficiency, including editing speed and number of learnable parameters.}
  \label{tab:efficiency}%
\end{table}%

\begin{figure}[t]
\setlength{\belowcaptionskip}{-0.3cm}
    \centering
    \includegraphics[width=0.95\linewidth]{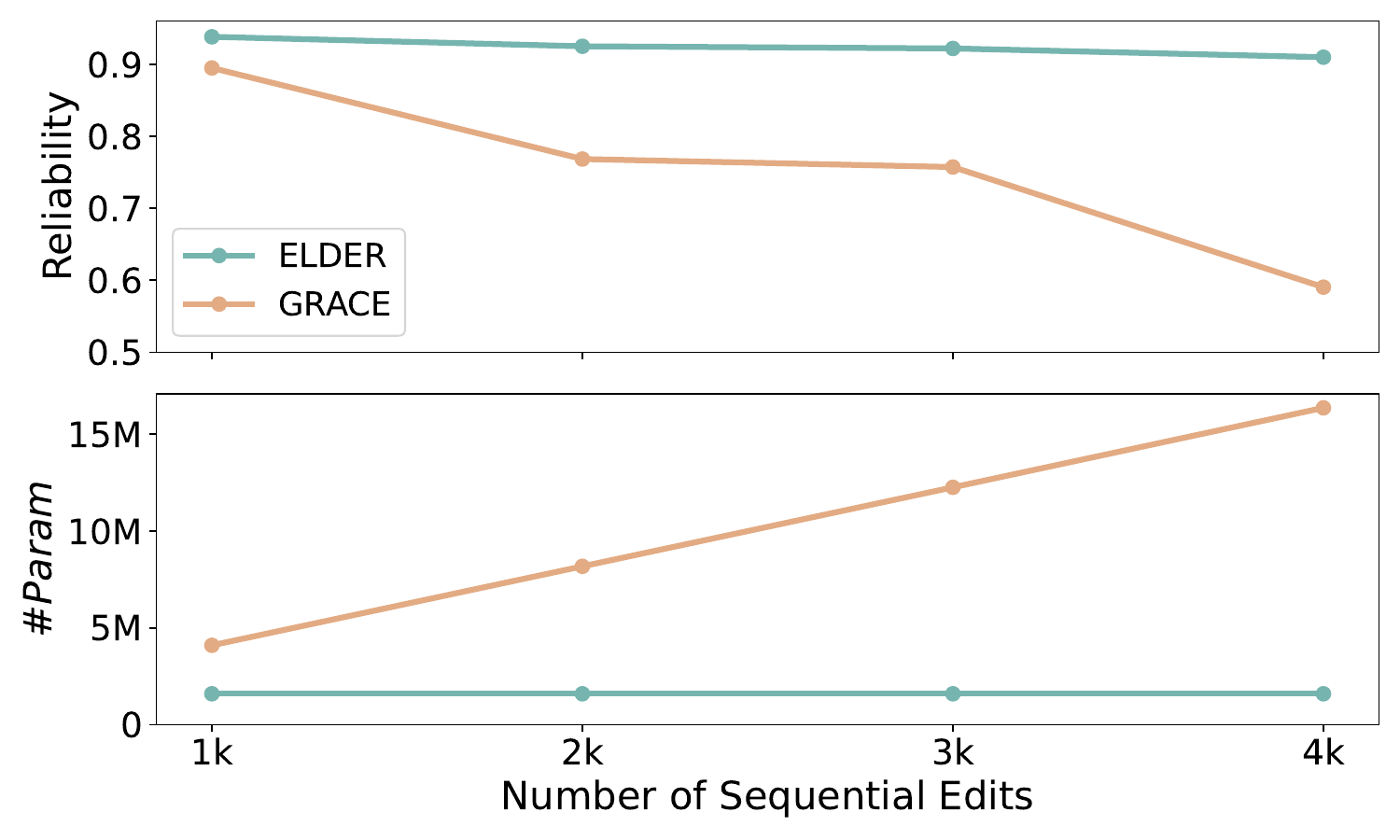}
    \caption{Editing scalability of GRACE and ELDER.}
    \label{fig:change_sequence_length_grace}
\end{figure}

\begin{figure}
    \centering
    \setlength{\belowcaptionskip}{-0.5cm}
    \includegraphics[width=0.95\linewidth]{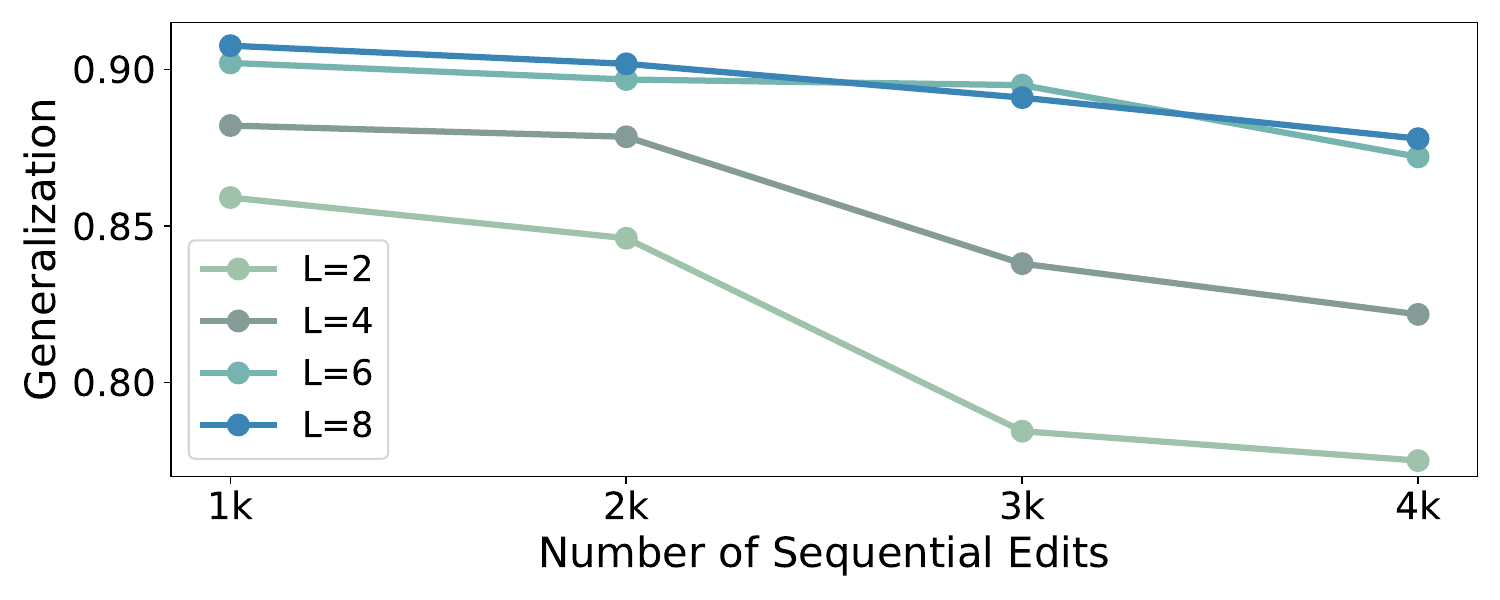}
    \caption{ELDER editing performance after varying numbers of edits with different parameter budgets.}
    \label{fig:change_L}
\end{figure}

We also examine how ELDER scales with different parameter budgets. 
Figure \ref{fig:change_L} presents the results of varying \(L\), which represents the number of layers utilizing the mixture-of-LoRA module, initially set to \(6\). 
As \(L\) increases, ELDER becomes more stable and exhibits better scalability, benefiting from the increased number of learnable parameters. 

\subsubsection{Ablation Experiments}\label{ablation}
We conduct ablation experiments to evaluate the effect of our proposed guided loss.
Two alternatives are considered: removing the guided loss during training or replacing it with the load balancing loss proposed by \citet{fedus2021switch}. 
The latter improves the routing scheme by calculating the average LoRA usage load per training batch and optimizes it to be uniformly distributed. 
The results, shown in Table\ref{tab:guided_loss}, demonstrate that the guided loss achieves superior results and enhances the training of the mixture-of-LoRAs for lifelong model editing.  
In contrast, directly uniforming usage load per batch degrades the performance, since the small batch size in lifelong editing leads to the disparity between batch measurement and real usage loads, as described in the Method Section\ref{sec: guide}.

\begin{table}[t]
\renewcommand{\arraystretch}{0.65} 
\setlength\tabcolsep{4pt}
  \centering
  \setlength{\belowcaptionskip}{-0.5cm}
    \begin{tabular}{lccccccc}
    \toprule
         & \multicolumn{3}{c}{ZsRE} &      & \multicolumn{3}{c}{\uppercase {C}\textsc{ounter}\uppercase {F}\textsc{act}} \\
\cmidrule{2-4}\cmidrule{6-8}         & GPT2&      & LLaMA2 &      & GPT2&      & LLaMA2 \\
    \midrule
    ELDER & 96.08  &      & 90.21  &      & 91.26  &      & 90.79  \\
    \textit{w/o guide} & 92.49  &      & 87.19  &      & 87.57  &      & 85.20  \\
    \textit{w balancing} & 74.26  &      & 71.39  &      & 75.49  &      & 77.94  \\
    \bottomrule
    \end{tabular}%
\caption{Editing generalization while altering the auxiliary loss. \textit{w/o guide} denotes removing guided loss from training. \textit{w balancing} means balancing load with the loss proposed by \citet{fedus2021switch} instead.}
\label{tab:guided_loss}
\end{table}%

\subsubsection{Model Analysis} \label{analysis}
\begin{figure}[ht]
     \centering
     \setlength{\belowcaptionskip}{-0.6cm}
     \includegraphics[width=0.82\linewidth]{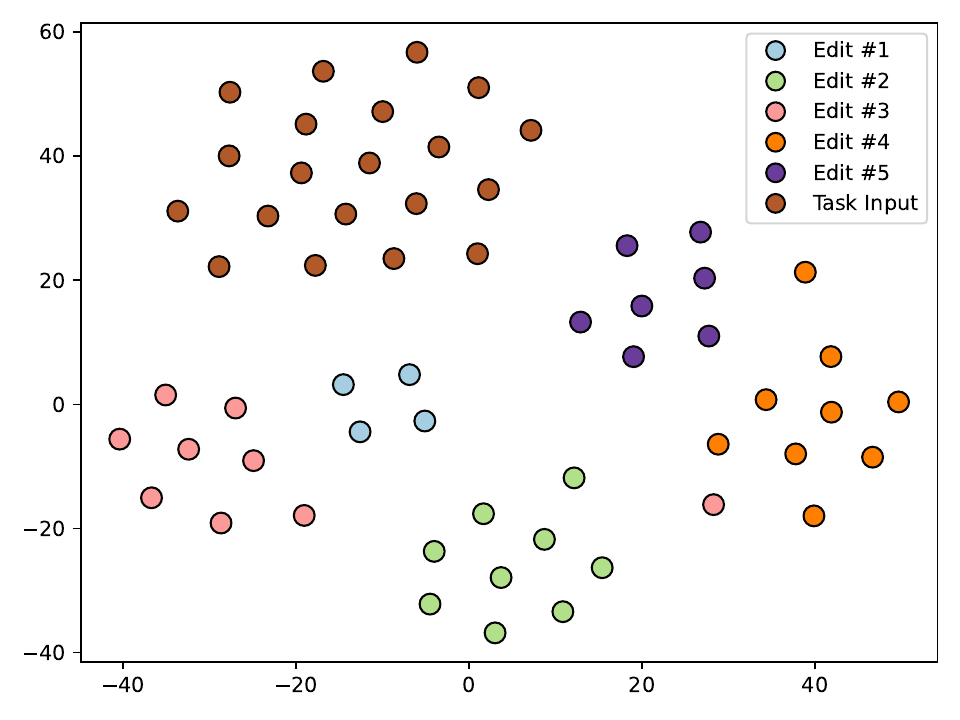}
     \caption{Visualization of LoRA allocation codes. Edit \#1 to \#5 denote five groups of semantically equivalent inputs, i.e., edits and their rephrases.}
     \label{fig:visualize}
 \end{figure}

We further assess the behavior of ELDER by visualizing the LoRA allocation codes generated by the mixture-of-LoRA modules. 
We use ZsRE and LLaMA2-7B for the analysis.
We randomly sample five groups of semantically equivalent inputs, including edits and their corresponding rephrases, along with twenty samples from the Natural Question dataset, chosen as task inputs unrelated to the edited knowledge.  
The allocation codes for these inputs are recorded and visualized using t-SNE, as shown in Figure \ref{fig:visualize}. 
We observe that semantically equivalent inputs receive similar, though not identical, LoRA allocations. 
This similarity suggests that ELDER effectively captures the semantic associations between edits and their rephrases, showing robustness to minor input variations. 
Additionally, task inputs are separated from all edits, supporting the use of allocation codes for discrimination in our deferral mechanism.

\section{Conclusion}

In conclusion, this paper presents ELDER, a novel approach for lifelong model editing using a mixture-of-LoRA structure. 
The training process is guided by a novel loss function to align the LoRA allocation with edit knowledge, and a deferral mechanism is integrated to retain general abilities after editing.
ELDER improves editing generalization by effectively assigning similar LoRA allocations to semantically equivalent inputs during inference.
Extensive experiments demonstrate that this method significantly enhances editing performance with remarkable efficiency and scalability, while also preserving the original performance of LLMs on general tasks.

\section{Acknowledgements}
We would like to thank Rui-chen Zheng for the discussion on optimizing adapter allocation.
This research is supported by Artificial Intelligence-National Science and Technology Major Project 2023ZD0121200 and  National Natural Science Foundation of China under Grant No.62222212, No.62376033, No.62372006.

\bibliography{aaai25}

\begin{thebibliography}{44}
\providecommand{\natexlab}[1]{#1}

\bibitem[{Achiam et~al.(2023)Achiam, Adler, Agarwal, Ahmad, Akkaya, Aleman, Almeida, Altenschmidt, Altman, Anadkat et~al.}]{gpt4}
Achiam, J.; Adler, S.; Agarwal, S.; Ahmad, L.; Akkaya, I.; Aleman, F.~L.; Almeida, D.; Altenschmidt, J.; Altman, S.; Anadkat, S.; et~al. 2023.
\newblock Gpt-4 technical report.
\newblock \emph{arXiv preprint arXiv:2303.08774}.

\bibitem[{Brown et~al.(2020)Brown, Mann, Ryder, Subbiah, Kaplan, Dhariwal, Neelakantan, Shyam, Sastry, Askell, Agarwal, Herbert-Voss, Krueger, Henighan, Child, Ramesh, Ziegler, Wu, Winter, Hesse, Chen, Sigler, Litwin, Gray, Chess, Clark, Berner, McCandlish, Radford, Sutskever, and Amodei}]{gpt3}
Brown, T.; Mann, B.; Ryder, N.; Subbiah, M.; Kaplan, J.~D.; Dhariwal, P.; Neelakantan, A.; Shyam, P.; Sastry, G.; Askell, A.; Agarwal, S.; Herbert-Voss, A.; Krueger, G.; Henighan, T.; Child, R.; Ramesh, A.; Ziegler, D.; Wu, J.; Winter, C.; Hesse, C.; Chen, M.; Sigler, E.; Litwin, M.; Gray, S.; Chess, B.; Clark, J.; Berner, C.; McCandlish, S.; Radford, A.; Sutskever, I.; and Amodei, D. 2020.
\newblock Language Models are Few-Shot Learners.
\newblock In Larochelle, H.; Ranzato, M.; Hadsell, R.; Balcan, M.; and Lin, H., eds., \emph{Advances in Neural Information Processing Systems}, volume~33, 1877--1901. Curran Associates, Inc.

\bibitem[{Clark et~al.(2019)Clark, Lee, Chang, Kwiatkowski, Collins, and Toutanova}]{clark2019boolq}
Clark, C.; Lee, K.; Chang, M.-W.; Kwiatkowski, T.; Collins, M.; and Toutanova, K. 2019.
\newblock BoolQ: Exploring the surprising difficulty of natural yes/no questions.
\newblock \emph{arXiv preprint arXiv:1905.10044}.

\bibitem[{Cobbe et~al.(2021)Cobbe, Kosaraju, Bavarian, Chen, Jun, Kaiser, Plappert, Tworek, Hilton, Nakano et~al.}]{cobbe2021training_gsm8k}
Cobbe, K.; Kosaraju, V.; Bavarian, M.; Chen, M.; Jun, H.; Kaiser, L.; Plappert, M.; Tworek, J.; Hilton, J.; Nakano, R.; et~al. 2021.
\newblock Training verifiers to solve math word problems.
\newblock \emph{arXiv preprint arXiv:2110.14168}.

\bibitem[{Cui et~al.(2020)Cui, Wu, Liu, Zhang, and Zhou}]{cui2020mutual}
Cui, L.; Wu, Y.; Liu, S.; Zhang, Y.; and Zhou, M. 2020.
\newblock MuTual: A dataset for multi-turn dialogue reasoning.
\newblock \emph{arXiv preprint arXiv:2004.04494}.

\bibitem[{Dagan, Glickman, and Magnini(2005)}]{dagan2005_rte}
Dagan, I.; Glickman, O.; and Magnini, B. 2005.
\newblock The pascal recognising textual entailment challenge.
\newblock In \emph{Machine learning challenges workshop}, 177--190. Springer.

\bibitem[{Dai et~al.(2022{\natexlab{a}})Dai, Dong, Hao, Sui, Chang, and Wei}]{dai2022knowledge_kn}
Dai, D.; Dong, L.; Hao, Y.; Sui, Z.; Chang, B.; and Wei, F. 2022{\natexlab{a}}.
\newblock Knowledge Neurons in Pretrained Transformers.
\newblock In \emph{Proceedings of the 60th Annual Meeting of the Association for Computational Linguistics (Volume 1: Long Papers)}, 8493--8502.

\bibitem[{Dai et~al.(2022{\natexlab{b}})Dai, Dong, Ma, Zheng, Sui, Chang, and Wei}]{dai2022stablemoe}
Dai, D.; Dong, L.; Ma, S.; Zheng, B.; Sui, Z.; Chang, B.; and Wei, F. 2022{\natexlab{b}}.
\newblock StableMoE: Stable Routing Strategy for Mixture of Experts.
\newblock In \emph{Proceedings of the 60th Annual Meeting of the Association for Computational Linguistics (Volume 1: Long Papers)}, 7085--7095.

\bibitem[{De~Cao, Aziz, and Titov(2021)}]{de2021editing_ke}
De~Cao, N.; Aziz, W.; and Titov, I. 2021.
\newblock Editing Factual Knowledge in Language Models.
\newblock In \emph{Proceedings of the 2021 Conference on Empirical Methods in Natural Language Processing}, 6491--6506.

\bibitem[{Dou et~al.(2023)Dou, Zhou, Liu, Gao, Zhao, Shen, Zhou, Xi, Wang, Fan et~al.}]{dou2023loramoe_2}
Dou, S.; Zhou, E.; Liu, Y.; Gao, S.; Zhao, J.; Shen, W.; Zhou, Y.; Xi, Z.; Wang, X.; Fan, X.; et~al. 2023.
\newblock Loramoe: Revolutionizing mixture of experts for maintaining world knowledge in language model alignment.
\newblock \emph{arXiv preprint arXiv:2312.09979}.

\bibitem[{Fedus, Zoph, and Shazeer(2021)}]{fedus2021switch}
Fedus, W.; Zoph, B.; and Shazeer, N. 2021.
\newblock Switch Transformers: Scaling to Trillion Parameter Models with Simple and Efficient Sparsity.
\newblock \emph{arXiv preprint arXiv:2101.03961}.

\bibitem[{Gu et~al.(2024)Gu, Xu, Ma, Lu, Ling, Chang, and Peng}]{gu2024model_hurt}
Gu, J.-C.; Xu, H.-X.; Ma, J.-Y.; Lu, P.; Ling, Z.-H.; Chang, K.-W.; and Peng, N. 2024.
\newblock Model editing can hurt general abilities of large language models.
\newblock \emph{arXiv preprint arXiv:2401.04700}.

\bibitem[{Gupta, Rao, and Anumanchipalli(2024)}]{gupta2024model}
Gupta, A.; Rao, A.; and Anumanchipalli, G. 2024.
\newblock Model Editing at Scale leads to Gradual and Catastrophic Forgetting.
\newblock \emph{arXiv preprint arXiv:2401.07453}.

\bibitem[{Hartvigsen et~al.(2024)Hartvigsen, Sankaranarayanan, Palangi, Kim, and Ghassemi}]{hartvigsen2024aging}
Hartvigsen, T.; Sankaranarayanan, S.; Palangi, H.; Kim, Y.; and Ghassemi, M. 2024.
\newblock Aging with grace: Lifelong model editing with discrete key-value adaptors.
\newblock \emph{Advances in Neural Information Processing Systems}, 36.

\bibitem[{Hu et~al.(2024)Hu, Cao, Chen, Liu, and Zhao}]{hu2024wilke}
Hu, C.; Cao, P.; Chen, Y.; Liu, K.; and Zhao, J. 2024.
\newblock WilKE: Wise-Layer Knowledge Editor for Lifelong Knowledge Editing.
\newblock \emph{arXiv preprint arXiv:2402.10987}.

\bibitem[{Hu et~al.(2021)Hu, Wallis, Allen-Zhu, Li, Wang, Wang, Chen et~al.}]{hu2021lora}
Hu, E.~J.; Wallis, P.; Allen-Zhu, Z.; Li, Y.; Wang, S.; Wang, L.; Chen, W.; et~al. 2021.
\newblock LoRA: Low-Rank Adaptation of Large Language Models.
\newblock In \emph{International Conference on Learning Representations}.

\bibitem[{Huang et~al.(2023)Huang, Liu, Lin, Pang, Du, and Lin}]{huang2023lorahub}
Huang, C.; Liu, Q.; Lin, B.~Y.; Pang, T.; Du, C.; and Lin, M. 2023.
\newblock Lorahub: Efficient cross-task generalization via dynamic lora composition.
\newblock \emph{arXiv preprint arXiv:2307.13269}.

\bibitem[{Huang et~al.(2022)Huang, Shen, Zhang, Zhou, Rong, and Xiong}]{huang2022transformer_patcher}
Huang, Z.; Shen, Y.; Zhang, X.; Zhou, J.; Rong, W.; and Xiong, Z. 2022.
\newblock Transformer-Patcher: One Mistake Worth One Neuron.
\newblock In \emph{The Eleventh International Conference on Learning Representations}.

\bibitem[{Jacobs et~al.(1991)Jacobs, Jordan, Nowlan, and Hinton}]{jacobs1991adaptive_moe_1}
Jacobs, R.~A.; Jordan, M.~I.; Nowlan, S.~J.; and Hinton, G.~E. 1991.
\newblock Adaptive mixtures of local experts.
\newblock \emph{Neural computation}, 3(1): 79--87.

\bibitem[{Ji et~al.(2023)Ji, Yu, Xu, Lee, Ishii, and Fung}]{ji2023towards}
Ji, Z.; Yu, T.; Xu, Y.; Lee, N.; Ishii, E.; and Fung, P. 2023.
\newblock Towards Mitigating LLM Hallucination via Self Reflection.
\newblock In \emph{Findings of the Association for Computational Linguistics: EMNLP 2023}, 1234--1245. Singapore: Association for Computational Linguistics.

\bibitem[{Jordan and Jacobs(1994)}]{jordan1994hierarchical_moe_2}
Jordan, M.~I.; and Jacobs, R.~A. 1994.
\newblock Hierarchical mixtures of experts and the EM algorithm.
\newblock \emph{Neural computation}, 6(2): 181--214.

\bibitem[{Kingma and Ba(2014)}]{kingma2014adam}
Kingma, D.~P.; and Ba, J. 2014.
\newblock Adam: A method for stochastic optimization.
\newblock \emph{arXiv preprint arXiv:1412.6980}.

\bibitem[{Kwiatkowski et~al.(2019)Kwiatkowski, Palomaki, Redfield, Collins, Parikh, Alberti, Epstein, Polosukhin, Devlin, Lee et~al.}]{kwiatkowski2019natural_NQ}
Kwiatkowski, T.; Palomaki, J.; Redfield, O.; Collins, M.; Parikh, A.; Alberti, C.; Epstein, D.; Polosukhin, I.; Devlin, J.; Lee, K.; et~al. 2019.
\newblock Natural questions: a benchmark for question answering research.
\newblock \emph{Transactions of the Association for Computational Linguistics}, 7: 453--466.

\bibitem[{Levy et~al.(2017)Levy, Seo, Choi, and Zettlemoyer}]{levy2017zsre}
Levy, O.; Seo, M.; Choi, E.; and Zettlemoyer, L. 2017.
\newblock Zero-shot relation extraction via reading comprehension.
\newblock \emph{arXiv preprint arXiv:1706.04115}.

\bibitem[{Lewis, Yao, and Ruder(2021)}]{lewis2021base}
Lewis, M.; Yao, Z.; and Ruder, S. 2021.
\newblock BASE Layers: Simplifying Training of Large, Sparse Models.
\newblock In \emph{Proceedings of the 38th International Conference on Machine Learning (ICML)}.

\bibitem[{Lin et~al.(2024)Lin, Beigi, Li, Zhou, Zhang, Wang, Yin, and Huang}]{lin2024navigating}
Lin, Z.; Beigi, M.; Li, H.; Zhou, Y.; Zhang, Y.; Wang, Q.; Yin, W.; and Huang, L. 2024.
\newblock Navigating the Dual Facets: A Comprehensive Evaluation of Sequential Memory Editing in Large Language Models.
\newblock \emph{arXiv preprint arXiv:2402.11122}.

\bibitem[{Meng et~al.(2022)Meng, Bau, Andonian, and Belinkov}]{meng2022locating_rome}
Meng, K.; Bau, D.; Andonian, A.; and Belinkov, Y. 2022.
\newblock Locating and editing factual associations in GPT.
\newblock \emph{Advances in Neural Information Processing Systems}, 35: 17359--17372.

\bibitem[{Mitchell et~al.(2021)Mitchell, Lin, Bosselut, Finn, and Manning}]{mitchell2021fast_mend}
Mitchell, E.; Lin, C.; Bosselut, A.; Finn, C.; and Manning, C.~D. 2021.
\newblock Fast Model Editing at Scale.
\newblock In \emph{International Conference on Learning Representations}.

\bibitem[{Mitchell et~al.(2022)Mitchell, Lin, Bosselut, Manning, and Finn}]{mitchell2022memory_serac}
Mitchell, E.; Lin, C.; Bosselut, A.; Manning, C.~D.; and Finn, C. 2022.
\newblock Memory-based model editing at scale.
\newblock In \emph{International Conference on Machine Learning}, 15817--15831. PMLR.

\bibitem[{Radford et~al.(2019)Radford, Wu, Child, Luan, Amodei, and Sutskever}]{radford2019gpt2}
Radford, A.; Wu, J.; Child, R.; Luan, D.; Amodei, D.; and Sutskever, I. 2019.
\newblock Language Models are Unsupervised Multitask Learners.
\newblock \emph{OpenAI blog}, 1(8): 9.

\bibitem[{Roller et~al.(2021)Roller, Sukhbaatar, Weston et~al.}]{roller2021hash}
Roller, S.; Sukhbaatar, S.; Weston, J.; et~al. 2021.
\newblock Hash layers for large sparse models.
\newblock \emph{Advances in Neural Information Processing Systems}, 34: 17555--17566.

\bibitem[{Sang and De~Meulder(2003)}]{sang2003conll}
Sang, E.~F.; and De~Meulder, F. 2003.
\newblock Introduction to the CoNLL-2003 shared task: Language-independent named entity recognition.
\newblock \emph{arXiv preprint cs/0306050}.

\bibitem[{Shazeer et~al.(2016)Shazeer, Mirhoseini, Maziarz, Davis, Le, Hinton, and Dean}]{shazeer2016outrageously_moe_3}
Shazeer, N.; Mirhoseini, A.; Maziarz, K.; Davis, A.; Le, Q.; Hinton, G.; and Dean, J. 2016.
\newblock Outrageously Large Neural Networks: The Sparsely-Gated Mixture-of-Experts Layer.
\newblock In \emph{International Conference on Learning Representations}.

\bibitem[{Sinitsin et~al.(2019)Sinitsin, Plokhotnyuk, Pyrkin, Popov, and Babenko}]{sinitsin2019editable}
Sinitsin, A.; Plokhotnyuk, V.; Pyrkin, D.; Popov, S.; and Babenko, A. 2019.
\newblock Editable Neural Networks.
\newblock In \emph{International Conference on Learning Representations}.

\bibitem[{Socher et~al.(2013)Socher, Perelygin, Wu, Chuang, Manning, Ng, and Potts}]{socher2013sst}
Socher, R.; Perelygin, A.; Wu, J.; Chuang, J.; Manning, C.~D.; Ng, A.~Y.; and Potts, C. 2013.
\newblock Recursive deep models for semantic compositionality over a sentiment treebank.
\newblock In \emph{Proceedings of the 2013 conference on empirical methods in natural language processing}, 1631--1642.

\bibitem[{Tam et~al.(2023)Tam, Mascarenhas, Zhang, Kwan, Bansal, and Raffel}]{tam2023evaluating}
Tam, D.; Mascarenhas, A.; Zhang, S.; Kwan, S.; Bansal, M.; and Raffel, C. 2023.
\newblock Evaluating the Factual Consistency of Large Language Models Through News Summarization.
\newblock In \emph{Findings of the Association for Computational Linguistics: ACL 2023}, 3456--3467. Toronto, Canada: Association for Computational Linguistics.

\bibitem[{Tian et~al.(2024)Tian, Cheng, Liang, Zhang, Hu, Xue, Gou, Chen, and Chen}]{tian2024instructedit}
Tian, B.; Cheng, S.; Liang, X.; Zhang, N.; Hu, Y.; Xue, K.; Gou, Y.; Chen, X.; and Chen, H. 2024.
\newblock InstructEdit: Instruction-based Knowledge Editing for Large Language Models.
\newblock \emph{arXiv preprint arXiv:2402.16123}.

\bibitem[{Touvron et~al.(2023)Touvron, Lavril, Izacard, Martinet, Lachaux, Goyal, Balland, Cucurull, Guestrin, Joulin et~al.}]{touvron2023llama2}
Touvron, H.; Lavril, T.; Izacard, G.; Martinet, X.; Lachaux, M.-A.; Goyal, N.; Balland, M.; Cucurull, G.; Guestrin, C.; Joulin, A.; et~al. 2023.
\newblock Llama 2: Open Foundation and Fine-Tuned Chat Models.
\newblock \emph{arXiv preprint arXiv:2307.09288}.

\bibitem[{Wang et~al.(2023)Wang, Yan, Huang, Zheng, and Huang}]{wang2023hallucination}
Wang, X.; Yan, Y.; Huang, L.; Zheng, X.; and Huang, X. 2023.
\newblock Hallucination Detection for Generative Large Language Models by Bayesian Sequential Estimation.
\newblock In \emph{Proceedings of the 2023 Conference on Empirical Methods in Natural Language Processing}, 15361--15371. Singapore: Association for Computational Linguistics.

\bibitem[{Yang et~al.(2024{\natexlab{a}})Yang, Ali, Wang, Hu, and Wang}]{yang2024moralmoeaugmentedlora}
Yang, S.; Ali, M.~A.; Wang, C.-L.; Hu, L.; and Wang, D. 2024{\natexlab{a}}.
\newblock MoRAL: MoE Augmented LoRA for LLMs' Lifelong Learning.
\newblock arXiv:2402.11260.

\bibitem[{Yang et~al.(2024{\natexlab{b}})Yang, Sun, Ma, Liu, Yin, and Cheng}]{yang2024butterfly}
Yang, W.; Sun, F.; Ma, X.; Liu, X.; Yin, D.; and Cheng, X. 2024{\natexlab{b}}.
\newblock The Butterfly Effect of Model Editing: Few Edits Can Trigger Large Language Models Collapse.
\newblock \emph{arXiv preprint arXiv:2402.09656}.

\bibitem[{Yao et~al.(2023)Yao, Wang, Tian, Cheng, Li, Deng, Chen, and Zhang}]{yao2023editing}
Yao, Y.; Wang, P.; Tian, B.; Cheng, S.; Li, Z.; Deng, S.; Chen, H.; and Zhang, N. 2023.
\newblock Editing Large Language Models: Problems, Methods, and Opportunities.
\newblock In \emph{Proceedings of the 2023 Conference on Empirical Methods in Natural Language Processing}, 10222--10240.

\bibitem[{Yu et~al.(2024)Yu, Chen, Zhou, and He}]{yu2024melo}
Yu, L.; Chen, Q.; Zhou, J.; and He, L. 2024.
\newblock Melo: Enhancing model editing with neuron-indexed dynamic lora.
\newblock In \emph{Proceedings of the AAAI Conference on Artificial Intelligence}, volume~38, 19449--19457.

\bibitem[{Zadouri et~al.(2023)Zadouri, {\"U}st{\"u}n, Ahmadian, Ermis, Locatelli, and Hooker}]{zadouri2023pushing_loramoe_1}
Zadouri, T.; {\"U}st{\"u}n, A.; Ahmadian, A.; Ermis, B.; Locatelli, A.; and Hooker, S. 2023.
\newblock Pushing Mixture of Experts to the Limit: Extremely Parameter Efficient MoE for Instruction Tuning.
\newblock In \emph{The Twelfth International Conference on Learning Representations}.

\end{thebibliography}

\newpage

\section{Appendix}
\subsection{Training Details}
We provide more training details for our experiments here.
All methods are trained on a single 48GB NVIDIA A40, and editing efficiency is evaluated on the same device.
We train ELDER with the Adam optimizer \citep{kingma2014adam}. 

The default hyperparameter setting we use are as follows.
For GPT2-XL, we edit the fully-connected component from the thirteen-fifth to fortieth layers (from \(transformer.h.36.mlp.c\_fc\) to \(transformer.h.40.mlp.c\_fc\)). 
For LLaMA2-7B, we edit the fully-connected component from the twentieth to the twenty-fifth layers (from \(model.layers.21.mlp.down\_proj\) to \(model.layers.26.mlp.down\_proj\)).
On all experiment settings, the learning rate of training ELDER is set to \(1e-4\), the batch size is set to \(4\), and we train for \(50\) iterations. 

For hyperparameter tuning, we list the tried values of several hyperparameters in ELDER, in the Table~\ref{tab:hyperparameter}.
During the development of this paper, we used \textit{Generalization} as the criterion when deciding the better \(k\), \(N\), and \(\lambda\), and used \textit{Test Retention on General Tasks} when altering \(\epsilon\).

\begin{table}[htbp]
  \centering

    \begin{tabular}{l|l}
    \toprule
    \multicolumn{1}{c|}{Hyperparameters } & \multicolumn{1}{c}{Values} \\
    \midrule
    k    & \{1, 2, 3, 4, 5\} \\
    N    & \{2, 4, 8, 16\} \\
     \(\lambda\) & \{0.001, 0.01, 0.1, 1, 10\} \\
     \(\epsilon\) & \{0, 6, 12, 18, 24\} \\
    \bottomrule
    \end{tabular}%
    \caption{Hyperparameter Values}
    \label{tab:hyperparameter}%
    
\end{table}%

\subsection{Additional Dataset Descriptions}
\subsubsection{Preprocessing Model Editing Dataset}
\label{sec:data}
We choose ZsRE and \uppercase {C}\textsc{ounter}\uppercase {F}\textsc{act} for experiments and utilize them following the methodologies outlined in \citep{hartvigsen2024aging} and \citep{yu2024melo}.
We use the same data to evaluate all baselines and ELDER.
For ZsRE, we follow the previous data split and evaluate all models on the validation set.
For \uppercase {C}\textsc{ounter}\uppercase {F}\textsc{act}, we use the original data file.
For both datasets, we extract the first 1000 edits with at least one rephrased data sample for the main experiments.
We split the edits and their rephrasings into two groups, using the first group to edit the model sequentially and the latter group to evaluate the editing generalization.

\subsubsection{Metrics of Test Retention Dataset}
\label{test_retention_metrics}
We evaluate the post-edit model on eight downstream general tasks and compute the test retention.
Following the original benchmark by \citet{gu2024model_hurt}, their respective metrics are listed below.
\begin{itemize}
    \item \textbf{Reasoning} on GSM8K \citep{cobbe2021training_gsm8k}. The results are measured by the solve rate.
    \item \textbf{Natural Language Inference} on RTE\citep{dagan2005_rte}. The results are measured by two-way classification.
    \item \textbf{Open-domain QA} on Natural Question\citep{kwiatkowski2019natural_NQ}. The results are measured by accuracy with the reference answer.
    \item  \textbf{Closed-domain QA} on BoolQ\citep{clark2019boolq}. The results are measured by exact match.
    \item  \textbf{Dialogue} on MuTual\citep{cui2020mutual}. The results are measured by selecting one best-matched response from four available choices.
    \item \textbf{Summarization} on SAMSum \citep{cui2020mutual}. The results are measured by the average of ROUGE-1, ROUGE-2, and ROGUE-L.
    \item \textbf{Named Entity Recognition} on CoNLL03\citep{sang2003conll}. The results are measured by entity-level F1-score.
    \item \textbf{Sentiment Analysis} on SST2\citep{socher2013sst}. Results are measured by the accuracy of two-way classification.
\end{itemize}

\subsection{Inference Efficiency}
We demonstrate the inference efficiency of ELDER by reporting the inference latency.  Our results show that ELDER maintains a similar inference speed to the base model, indicating its effectiveness for practical usage. The statistics are measured after 1000 edits from the ZsRE dataset.
\begin{table}[htbp]
  \centering
  
    \begin{tabular}{llrl}
    \toprule
         & GPT2-XL &      & LLaMA2-7B \\
\cmidrule{2-4}    Base & 0.56 s/edit &      & 1.01 s/edit \\
    ELDER & 0.58 s/edit &      & 1.04 s/edit \\
    \bottomrule
    \end{tabular}%
  \label{tab:addlabel}%
  \caption{Inference Efficiency}
\end{table}%

\subsection{Detailed Results of Test Retention on General Tasks}
\label{sec:retention}

The detailed results of the pre-edit and post-edit models on downstream tasks under each experimental setting are shown in Table \ref{tab:zsre_llama}, Table \ref{tab:zsre_gpt}, Table \ref{tab:cf_llama} and Table \ref{tab:cf_gpt}. 
Each task has been described before.
The last column in each table stands for the standard \textit{Test Retention on General Tasks} we report in the main paper. 


\begin{table*}[htbp]
  \centering
  
    \begin{tabular}{lccccccccc}
    \toprule
    Method & RTE  & BoolQ & SST2 & MuTual & CoNLL03 & NQ   & GSM8K & SAMSum & \textit{Avg.} \\
    \midrule
    Base & 53.4  & 59.9  & 52.6  & 22.8  & 14.7  & 22.9  & 16.8  & 13.7  & 32.1  \\
    \midrule
    FT-L & 49.2  & 27.7  & 44.4  & 17.2  & 13.0  & 22.0  & 16.6  & 11.7  & 25.2  \\
    LoRA & 8.2  & 5.1  & 8.3  & 3.8  & 4.4  & 13.1  & 2.2  & 2.5  & 6.0  \\
    ROME & 14.1  & 16.6  & 0.6  & 1.0  & 1.5  & 6.8  & 7.1  & 5.9  & 6.7  \\
    WilKE & 22.4  & 8.9  & 3.6  & 6.8  & 5.2  & 9.6  & 3.9  & 1.8  & 7.8  \\
    SERAC & 54.1  & 59.9  & 52.6  & 22.3  & 14.7  & 22.6  & 16.2  & 13.7  & 32.0  \\
    T-Patcher & 0.0  & 0.2  & 0.0  & 0.5  & 0.3  & 6.2  & 0.0  & 1.3  & 1.1  \\
    MELO & 52.1  & 56.8  & 49.1  & 19.6  & 13.2  & 19.7  & 15.1  & 13.4  & 29.9  \\
    GRACE & 53.4  & 60.1  & 58.1  & 21.4  & 14.7  & 23.2  & 15.8  & 14.6  & 32.7  \\
    \midrule
    ELDER & 53.0  & 60.1  & 57.7  & 21.7  & 9.7  & 24.5  & 17.7  & 13.7  & 32.3  \\
    \bottomrule
    \end{tabular}%

  \caption{Results on general task of LLaMA2-7B after editing with ZsRE}
  \label{tab:zsre_llama}%
\end{table*}%

\begin{table*}[htbp]
  \centering
 
    \begin{tabular}{lccccccccc}
    \toprule
    Method & RTE  & BoolQ & SST2 & MuTual & CoNLL03 & NQ   & GSM8K & SAMSum & \textit{Avg.} \\
    \midrule
    Base & 51.8  & 51.9  & 48.9  & 19.9  & 18.7  & 34.2  & 4.0  & 12.5  & 30.2  \\
    \midrule
    FT-L & 0.0  & 0.0  & 0.0  & 2.9  & 1.2  & 0.7  & 0.0  & 1.1  & 0.7  \\
    LoRA & 0.0  & 0.0  & 0.0  & 0.2  & 1.7  & 0.0  & 0.0  & 0.0  & 0.2  \\
    ROME & 0.0  & 0.0  & 0.0  & 0.0  & 0.5  & 0.1  & 0.0  & 0.6  & 0.2  \\
    WilKE & 4.4  & 0.0  & 5.7  & 0.8  & 0.3  & 0.0  & 0.0  & 5.5  & 2.1  \\
    SERAC & 51.8  & 51.9  & 48.8  & 19.5  & 18.7  & 34.5  & 4.0  & 12.5  & 30.2  \\
    T-Patcher & 0.0  & 3.0  & 3.1  & 0.0  & 0.0  & 10.2  & 0.0  & 4.9  & 2.7  \\
    MELO & 51.3  & 49.5  & 47.2  & 20.9  & 18.1  & 33.2  & 4.1  & 12.1  & 29.6  \\
    GRACE & 51.8  & 51.9  & 48.9  & 19.9  & 18.7  & 34.2  & 4.0  & 12.5  & 30.3  \\
    \midrule
    ELDER & 51.6  & 52.0  & 47.5  & 21.0  & 18.7  & 33.4  & 4.3  & 12.5  & 30.1  \\
    \bottomrule
    \end{tabular}%
  
   \caption{Results on general task of GPT2-XL after editing with ZsRE}
   \label{tab:zsre_gpt}%
\end{table*}%

\begin{table*}[htbp]
  \centering
  
    \begin{tabular}{lccccccccc}
    \toprule
    Method & RTE  & BoolQ & SST2 & MuTual & CoNLL03 & NQ   & GSM8K & SAMSum & \textit{Avg.} \\
    \midrule
    Base & 53.4  & 59.9  & 52.6  & 22.8  & 14.7  & 22.9  & 16.8  & 13.7  & 32.1  \\
    \midrule
    FT-L & 3.9  & 0.0  & 0.1  & 7.5  & 0.6  & 22.0  & 16.6  & 11.7  & 7.8  \\
    LoRA & 0.2  & 0.0  & 0.8  & 0.3  & 0.0  & 3.7  & 0.0  & 0.0  & 0.6  \\
    ROME & 0.0  & 0.0  & 0.0  & 2.3  & 0.5  & 0.2  & 0.0  & 0.7  & 0.5  \\
    WilKE & 0.0  & 15.4  & 0.6  & 3.1  & 0.8  & 21.5  & 0.0  & 2.2  & 5.5  \\
    SERAC & 53.4  & 59.9  & 52.6  & 23.3  & 14.7  & 22.0  & 16.8  & 13.7  & 32.1  \\
    T-Patcher & 0.0  & 5.8  & 0.4  & 0.0  & 0.2  & 0.0  & 0.0  & 0.9  & 0.9  \\
    MELO & 52.9  & 58.1  & 53.2  & 25.4  & 14.8  & 21.2  & 17.4  & 12.5  & 31.9  \\
    GRACE & 53.1  & 59.8  & 54.7  & 21.2  & 14.9  & 22.6  & 16.8  & 14.4  & 32.2  \\
    \midrule
    ELDER & 53.3  & 59.9  & 50.1  & 22.3  & 14.2  & 22.9  & 17.0  & 11.9  & 31.5  \\
    \bottomrule
    \end{tabular}%
  
  \caption{Results on general task of LLaMA2-7B after editing with \uppercase {C}\textsc{ounter}\uppercase {F}\textsc{act}}
  \label{tab:cf_llama}%
\end{table*}%

\begin{table*}[htbp]
  \centering
    \begin{tabular}{lccccccccc}
    \toprule
    Method & RTE  & BoolQ & SST2 & MuTual & CoNLL03 & NQ   & GSM8K & SAMSum & \textit{Avg.} \\
    \midrule
    Base & 51.8  & 51.9  & 48.9  & 19.9  & 18.7  & 34.2  & 4.0  & 12.5  & 30.2  \\
    \midrule
    FT-L & 0.0  & 0.0  & 0.0  & 2.9  & 1.2  & 0.7  & 0.0  & 0.4  & 0.7  \\
    LoRA & 0.0  & 0.0  & 0.0  & 0.8  & 0.0  & 0.2  & 0.0  & 0.9  & 0.2  \\
    ROME & 0.0  & 0.0  & 0.0  & 0.0  & 0.6  & 0.3  & 0.0  & 0.5  & 0.2  \\
    WilKE & 0.0  & 0.0  & 0.0  & 4.2  & 4.5  & 0.9  & 0.0  & 0.0  & 1.2  \\
    SERAC & 51.8  & 52.4  & 49.6  & 21.4  & 18.7  & 34.2  & 6.4  & 12.5  & 30.9  \\
    T-Patcher & 0.0  & 0.0  & 0.0  & 0.0  & 0.8  & 0.2  & 0.0  & 0.0  & 0.1  \\
    MELO & 52.5  & 51.3  & 51.1  & 24.1  & 17.9  & 32.2  & 8.1  & 13.7  & 31.4  \\
    GRACE & 51.8  & 51.9  & 48.9  & 19.9  & 18.7  & 34.2  & 4.0  & 12.5  & 30.3  \\
    \midrule
    ELDER & 51.8  & 51.9  & 48.8  & 22.5  & 18.5  & 34.2  & 4.2  & 11.9  & 30.5  \\
    \bottomrule
    \end{tabular}%

    \caption{Results on general task of GPT2-XL after editing with \uppercase {C}\textsc{ounter}\uppercase {F}\textsc{act}}
      \label{tab:cf_gpt}%
\end{table*}%

\end{document}